\documentclass[conference]{IEEEtran}
\usepackage{times}

\usepackage[numbers]{natbib}
\usepackage{multicol}
\usepackage[bookmarks=true]{hyperref}
\usepackage{xcolor}
\usepackage{graphicx}
\usepackage{subcaption}

\begin{document}

\title{Domain Adaptation and Multi-view Attention for Learnable Landmark Tracking with Sparse Data}

\author{\authorblockN{Timothy Chase Jr, Karthik Dantu}
\authorblockA{Department of Computer Science and Engineering\\
University at Buffalo\\
Buffalo, New York 14260\\
Email: tbchase@buffalo.edu}}


%

\maketitle

\begin{abstract}
The detection and tracking of celestial surface terrain features are crucial for autonomous spaceflight applications, including Terrain Relative Navigation (TRN), Entry, Descent, and Landing (EDL), hazard analysis, and scientific data collection. Traditional photoclinometry-based pipelines often rely on extensive a priori imaging and offline processing, constrained by the computational limitations of radiation-hardened systems. While historically effective, these approaches typically increase mission costs and duration, operate at low processing rates, and have limited generalization. Recently, learning-based computer vision has gained popularity to enhance spacecraft autonomy and overcome these limitations. While promising, emerging techniques frequently impose computational demands exceeding the capabilities of typical spacecraft hardware for real-time operation and are further challenged by the scarcity of labeled training data for diverse extraterrestrial environments. In this work, we present novel formulations for in-situ landmark tracking via detection and description. We utilize lightweight, computationally efficient neural network architectures designed for real-time execution on current-generation spacecraft flight processors. For landmark detection, we propose improved domain adaptation methods that enable the identification of celestial terrain features with distinct, cheaply acquired training data. Concurrently, for landmark description, we introduce a novel attention alignment formulation that learns robust feature representations that maintain correspondence despite significant landmark viewpoint variations. Together, these contributions form a unified system for landmark tracking that demonstrates superior performance compared to existing state-of-the-art techniques. \href{https://tjchase34.github.io/assets/pdfs/rss_poster.pdf}{\textcolor{blue}{Poster available online.}}

\end{abstract}

\IEEEpeerreviewmaketitle

\section{Introduction}
The exploration of deep space objects necessitates complex autonomous spacecraft operations, such as TRN, EDL, hazard analysis, and scientific data collection, where communication latencies preclude real-time Earth-based guidance. Established photoclinometry-based pipelines proven on missions like Mars Perseverance~\cite{m2020lvs} and OSIRIS-REx~\cite{nft} employ template correlation techniques with patch-based landmarks identified on static navigation maps constructed \textit{a priori}~\cite{m2020lvs, nft, retina}, relying on feature (i.e., landmark) selection and offline processing by human operators. This approach leads to slow processing, high mission costs/durations, and poor generalization to unseen data. While recent space-grade neural network accelerators~\cite{sclearn} enable more state-of-the-art computer vision techniques capable of real-time performance~\cite{profiling}, achieving robust in-situ landmark tracking remains difficult in unstructured space environments~\cite{slamprob}.

Implementing such learning-based systems involves addressing key challenges. For landmark detection, reliance on supervised learning is impractical due to the scarcity of labeled celestial data. Unsupervised Domain Adaptation (UDA) provides an alternative, adapting models trained on source data (e.g., simulations) to unlabeled target domain imagery, though specific difficulties like illumination variance, feature space ambiguity, and textureless regions must be overcome. Visual Similarity-based Alignment (VSA) offers a promising UDA technique to address these issues. Originally developed for two-stage detectors~\cite{visga}, VSA mitigates distribution shifts by grouping object features based on visual appearance into instance-level clusters, thus reducing dependence on potentially noisy class and pseudo-class labels during training. 




Subsequently, robust landmark description remains an open problem. On Earth, related challenges are addressed through representation learning tasks, where the core objective is to learn discriminative embeddings via metric learning. Landmark recognition presents heightened complexity, as it requires distinguishing between members of a single geological subclass (e.g., all craters on the moon), a task arguably more fine-grained than current terrestrial benchmarks. Additionally, the visual appearance of any single landmark can vary dramatically between observations due to viewpoint changes, illumination shifts, and scale differences. Such challenges pose considerable difficulty for conventional approaches to effectively reason about landmark identity on their own.


\section{Methodology}

We introduce two distinct yet complementary components to a fully in-situ and learning based landmark tracking pipeline, namely You Only Crash Once (YOCO)~\cite{yoco2} for detection and Multi-view Attention Regularizations (MARs)~\cite{mars} for description. 

\subsection{Detection: You Only Crash Once (YOCO)}
\begin{figure}[h!]
    \centering
    \setkeys{Gin}{width=\columnwidth}
    \begin{subfigure}{0.48\columnwidth}
        \includegraphics{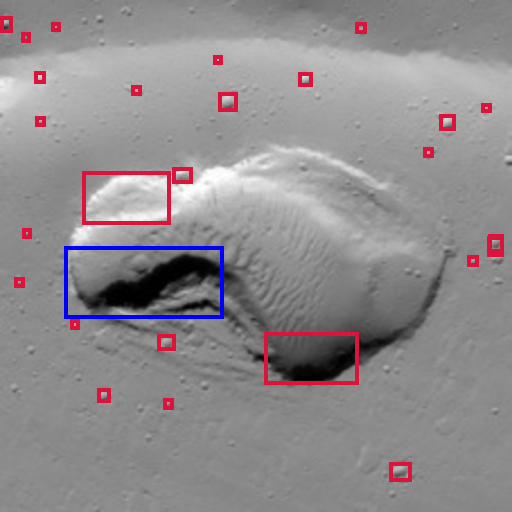}\\[3pt]
        \includegraphics{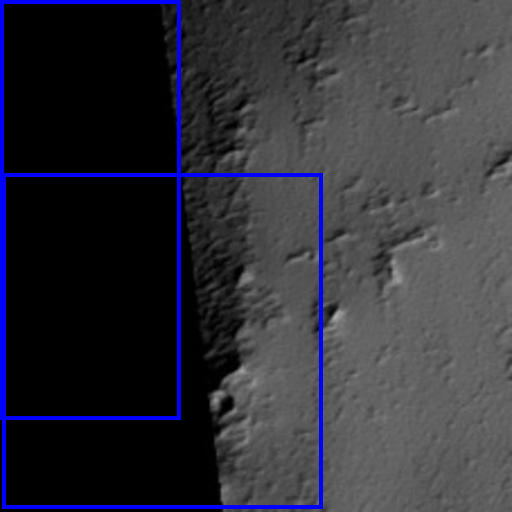}\\[3pt]
        \includegraphics{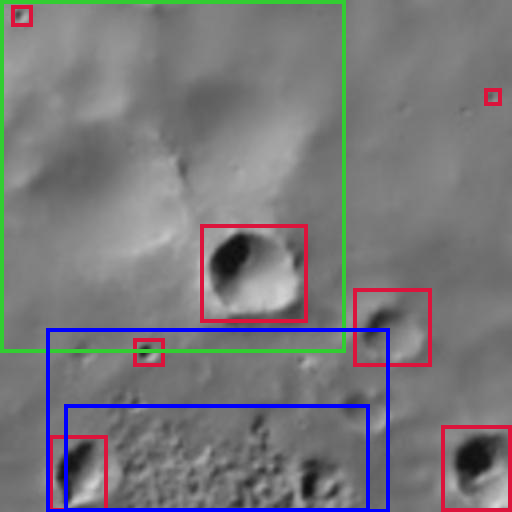}
        \caption{YOLO v5}
    \end{subfigure}
    \begin{subfigure}{0.48\columnwidth}
        \includegraphics{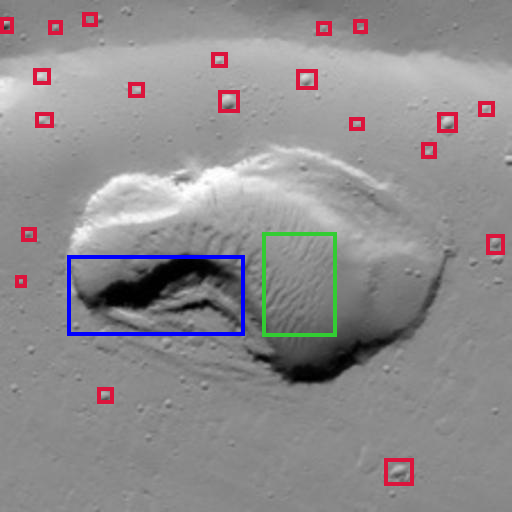}\\[3pt]
        \includegraphics{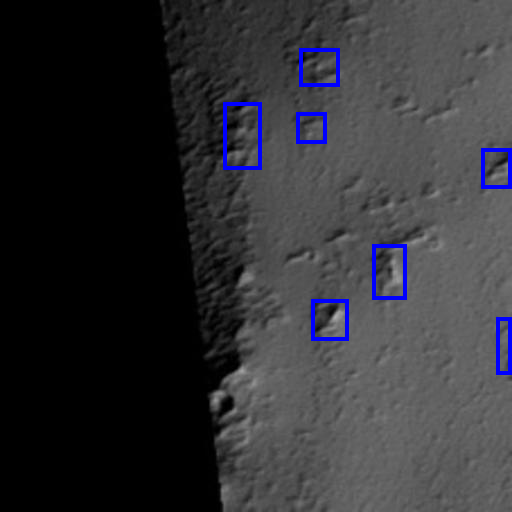}\\[3pt]
        \includegraphics{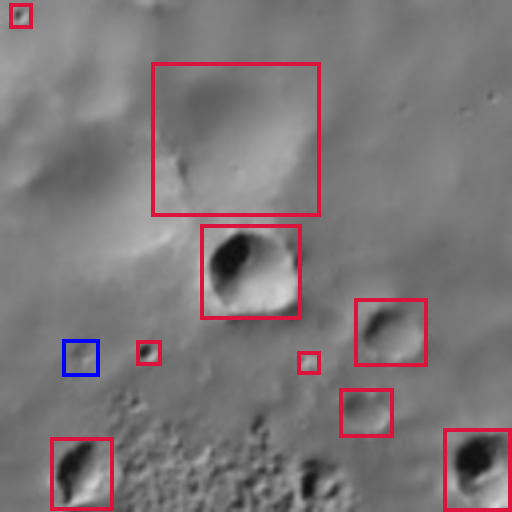}
        \caption{YOCO}
    \end{subfigure}
    \caption{YOLO (source-only) vs YOCO (proposed UDA) detection results on Mars HiRISE imagery. Red is \textit{crater}, blue is \textit{mountain}, green is \textit{sand dune}.}
    \label{fig:qual_mars}
\end{figure}

YOCO permits real-time and performant object detection on computationally constrained flight hardware (e.g., 79 ms inference on Zynq-7020, USB 2.0 Edge TPU). YOCO integrates UDA into one-stage YOLO architectures, enabling robust terrain feature detection using labeled source data (e.g., simulations) alongside unlabeled target imagery from challenging planetary, lunar, or small-body environments, specifically targeting improvements for textureless regions and varied illumination. The approach combines standard supervised loss with UDA terms, aligning global and local features. Local alignment first employs source-target feature regularization and unsupervised feature selection, driven by adversarial and contrastive VSA losses to minimize the domain gap. Quantitative and qualitative (\autoref{fig:qual_mars}) results on Mars data significantly outperform baseline models trained only on source data by drastically reducing false detections and improving accuracy.


\subsection{Description: Multi-view Attention Regularizations (MARs)}
\begin{figure}[h!]
    \centering
    \setkeys{Gin}{width=\columnwidth}
    \begin{subfigure}{0.48\columnwidth}
        \includegraphics{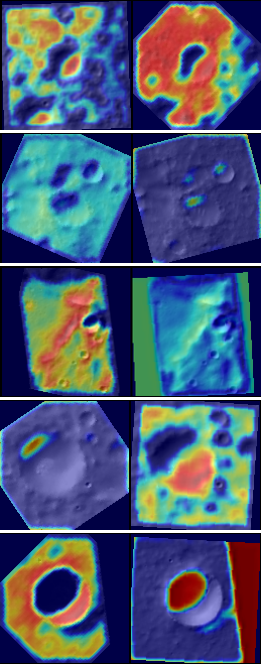}
        \caption{Traditional Attention Focus}
    \end{subfigure}
    \begin{subfigure}{0.48\columnwidth}
        \includegraphics{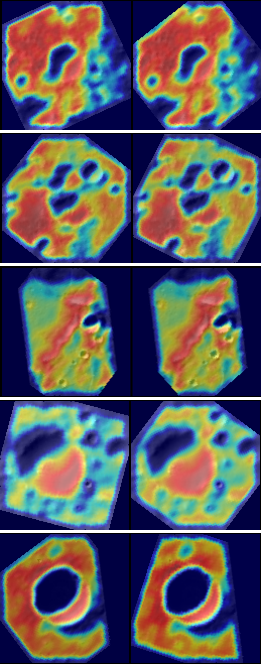}
        \caption{Attention Focus with MARs}
    \end{subfigure}
    \caption{Visualizations of attention focus on Mars craters with multiple views.}
    \label{fig:mars}
\end{figure}

The description component introduces Multi-view Attention Regularizations (MARs) to enhance metric learning for robust landmark recognition, specifically addressing performance issues linked to view-unaware attention mechanisms in standard architectures. MARs aims to improve landmark embedding distinguishability by making network attention implicitly view-aware, enforcing consistency in channel and spatial attention scaling factors. This constrains the focus of attention on the same features/regions across different views of the same landmark. Operating within a contrastive learning framework, MARs introduces auxiliary similarity losses applied to attention maps extracted from intermediate feature extraction stages. These maps are embedded into separate channel and spatial metric spaces, and the MARs constraint penalizes divergence between attention embeddings from positive view pairs. Validated across Earth, Mars, and Moon datasets (including a new photorealistic lunar dataset, Luna-1), MARs demonstrably enforces strong attention correlation between views (\autoref{fig:mars}), leading to significant improvements in overall recognition performance compared to baseline methods.

\section{Conclusion}
This research addresses the critical challenges hindering robust, real-time visual navigation for autonomous spacecraft using learning-based methods. Traditional systems are constrained by reliance on \textit{a priori} data and limited onboard processing, while standard deep learning techniques face hurdles related to computational cost and data label scarcity in extraterrestrial settings. We present novel contributions for both landmark detection (YOCO) and description (MARs). Our enhanced unsupervised domain adaptation framework enables accurate, real-time detection on flight hardware using cheaply acquired source and unlabeled target data, overcoming key environmental challenges such as texture sparsity and illumination variation. Complementarily, our proposed Multi-view Attention Regularizations significantly improve landmark description robustness by enforcing attention consistency across diverse viewpoints. This integrated approach demonstrably advances the state-of-the-art in visual processing, offering enhanced capabilities essential for enabling complex autonomous operations in future space exploration missions.


\bibliographystyle{plainnat}
\bibliography{references}

\end{document}